\pdfoutput=1

\documentclass[11pt]{article}

\usepackage[final]{acl}

\usepackage{times}
\usepackage{latexsym}

\usepackage[T1]{fontenc}

\usepackage[utf8]{inputenc}

\usepackage{microtype}

\usepackage{graphicx}
\usepackage{stfloats}

\usepackage{array}
\newcolumntype{C}[1]{>{\centering\let\newline\\\arraybackslash\hspace{0pt}}m{#1}}
  
\usepackage{courier}
\usepackage{multirow}
\usepackage{xcolor}
\usepackage{xargs}
\usepackage[colorinlistoftodos,prependcaption,textsize=tiny]{todonotes}
\usepackage{adjustbox}

\definecolor{DeepSkyBlue}{RGB}{0,191,255}
\newcommandx{\review}[2][1=]{\todo[linecolor=DeepSkyBlue,backgroundcolor=DeepSkyBlue!25,bordercolor=DeepSkyBlue,#1]{#2}}

\title{Logical Reasoning for Task Oriented Dialogue Systems}
\author{%
  Sajjad~Beygi \\
  \texttt{beygi@amazon.com} 
   \And
   Maryam~Fazel-Zarandi \\
  \texttt{fazelzar@amazon.com} 
   \AND
   Alessandra~Cervone \\
  \texttt{cervon@amazon.com} 
   \And
   Prakash~Krishnan \\
  \texttt{prakaskr@amazon.com} 
     \And
   Siddhartha~Reddy~Jonnalagadda \\
  \texttt{sjjonnal@amazon.com} 
}

\begin{document}
\maketitle
\vspace{-2.0in}
\begin{abstract}
In recent years, large pretrained models have been used in dialogue systems to improve successful task completion rates. However, lack of reasoning capabilities of dialogue platforms make it difficult to provide relevant and fluent responses, unless the designers of a conversational experience spend a considerable amount of time implementing these capabilities in external rule based modules. In this work, we propose a novel method to fine-tune pretrained transformer models such as Roberta and T5.  to reason over a set of facts in a given dialogue context.
Our method includes a synthetic data generation mechanism which helps the model learn logical relations, such as comparison between list of numerical values, inverse relations (and negation), inclusion and exclusion for categorical attributes, and application of a combination of attributes over both numerical and categorical values, and spoken form for numerical values, without need for additional training dataset.  We show that the transformer based model can perform logical reasoning to answer questions when the dialogue context contains all the required information, otherwise it is able to extract appropriate constraints to pass to downstream components (e.g. a knowledge base) when partial information is available. We observe that transformer based models such as UnifiedQA-T5 can be fine-tuned to perform logical reasoning (such as numerical and categorical attributes' comparison) over attributes that been seen in training time (e.g.,  accuracy of 90\%+ for comparison of smaller than $k_{\max}$=5 values over heldout test dataset). 

\end{abstract}

\section{Introduction}

\begin{figure}[t]
\centering
  \includegraphics[width=0.9\linewidth]{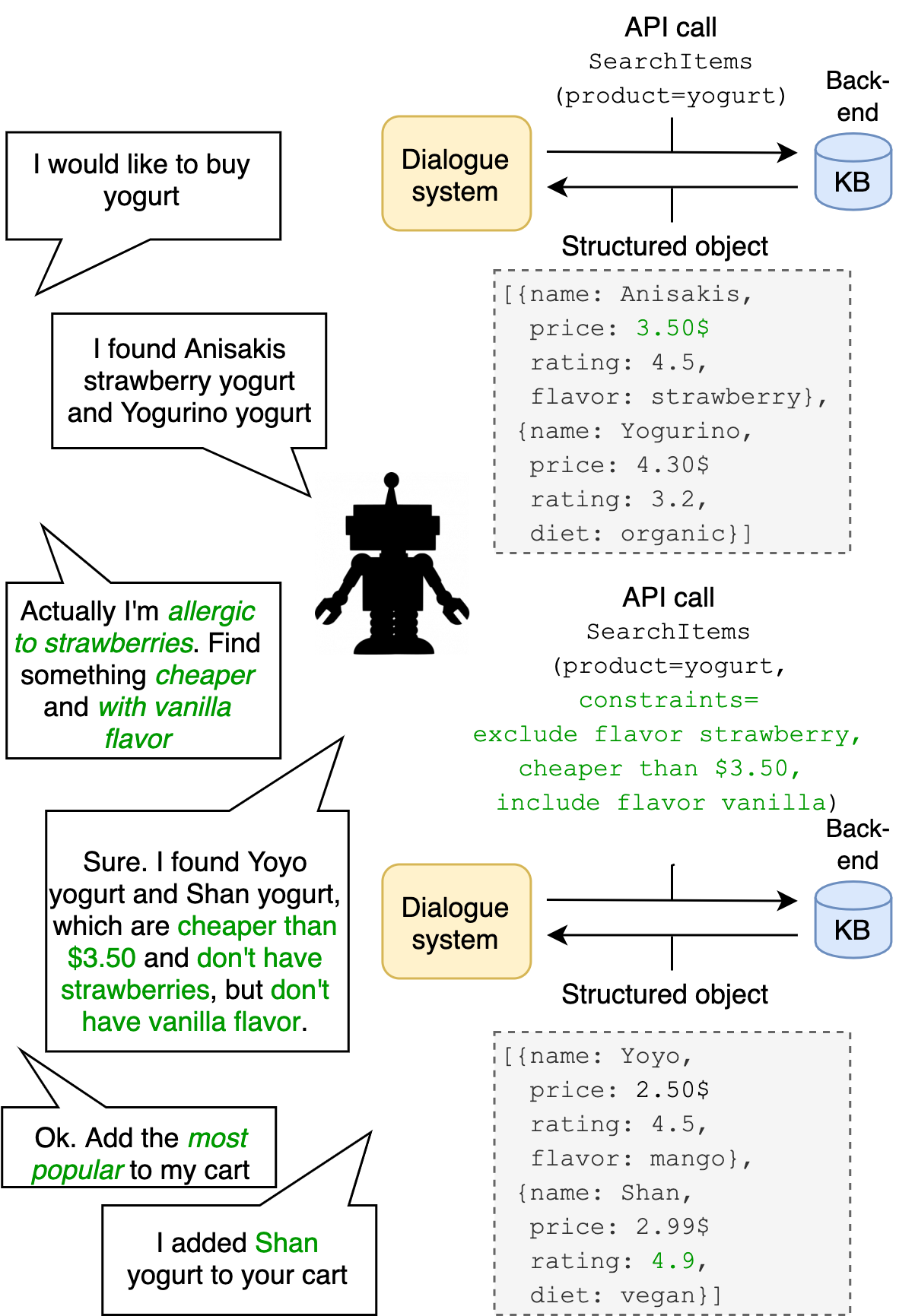}
  \caption{The dialogue system with reasoning ability.}
    \vspace{-15pt}
  \label{fig:figure_0001}
\end{figure}

Logical reasoning is an important aspect of human thinking and communication. 
Humans reason over beliefs, preferences, time, facts, and other contextual information to achieve complex tasks, derive meaning, and analyze emotions. 
Current task-oriented dialogue systems, however, only support very limited forms of logical reasoning. 
More specifically, although reasoning ability has been investigated as part of chatbots \cite{mutual} and question-answering systems \cite{huang2019cosmos, chen2020},
in many task-oriented dialogue systems today, the reasoning is mainly focused on determining which slot values 
are still unknown to the system but are required and elicit them \cite{guo2017qa}. 
However, in realistic task-oriented dialogues, logical reasoning is required to understand the user's request, 
ask questions that help address the user's task successfully and minimize asking irrelevant questions. 
The lack of robust, generalizable reasoning capabilities for dialogue systems, requires developers of the system to spend a considerable amount of time implementing these capabilities in external, rule-based and domain specific components.
This leads to a poor user experience requiring users to often correct the system's understanding, 
repeat themselves to ask the same question in different ways, restart the conversation when the system fails 
to recover from a `dead-end', or even change their goal. 

In this work, we propose to build on recent advances in research on logical reasoning and deep networks (e.g., 
\citeauthor{dong2019neural} \citeyear{dong2019neural}; \citeauthor{wang2019satnet} \citeyear{wang2019satnet}; \citeauthor{xie2019embedding} \citeyear{xie2019embedding}; \citeauthor{clark2020transformers} \citeyear{clark2020transformers}; \citeauthor{arabshahi2020conversational} \citeyear{arabshahi2020conversational}) 
to bring reasoning capabilities to task-oriented dialogue systems. 
Our primary focus in this work is on mechanisms by which logical reasoning can be learned and used in conversational systems. 
In this direction, we propose a novel deep learning method to fine-tune pretrained models to reason over numerical and categorical attributes in the dialogue context and present an architecture for the integration of this model in task-oriented dialogue systems.
Our objective is for the model to do logical reasoning to respond to queries from the dialogue context when it has all the required information available in the dialogue context without additional external logic (e.g., ``Add the most popular to my cart'' in Figure \ref{fig:figure_0001}), 
extract constraints and inform downstream components when it only has partial context (e.g., ``Actually I'm allergic to berries. Find something cheaper and with vanilla flavor'' in Figure \ref{fig:figure_0001}, where cheaper means cheaper than what was shown so far), 
and not provide an answer when it does not have any relevant information and delegate to the dialogue policy to determine the next action. 



We specifically choose to fine-tune transformers since these models operate on language directly, 
do not impose any structure on the reasoning process \cite{clark2020transformers}, 
and we can leverage the knowledge and diversity of language that the pretrained models have already learned. 
Furthermore, \citeauthor{ding2020object} (\citeyear{ding2020object}) recently showed that these approaches can outperform neuro-symbolic methods. 
Our approach is similar to recent works on using transformers as soft reasoners  \cite{clark2020transformers, talmor2020leap}. 
However, compared to these methods, we focus on use cases relevant to conversational systems and our model goes beyond predicting a 
true/false response to directly predicting the answer when the model has the information or 
extract constraints when it has partial information.
In this direction, we report experimental results that show using our training method
transformers can learn to reason over numerical and categorical attributes in the dialogue context. 


Note that although we use transformers for our experiments, our proposed method can be used to generate data and train 
any other seq2seq model for the same task and be integrated with any dialogue system in a similar manner. 
Furthermore, our proposed method is different from question-answering or machine reading comprehension in that we are not looking for an answer in a specific passage; 
rather, we want the model to reason over facts in the dialogue context to draw parallels and conclusions to inform decision making, 
similar to how humans reason over a multi-turn conversation. 

\section{Related Work}

The approaches for integrating reasoning with deep networks can be categorized into the following.
 

\paragraph{Reasoning after Semantic Parsing} These approaches convert utterances to a semantic representation and feed it to a set of rules or a formal reasoner for reasoning. 
For example, \citeauthor{kamath2018survey} (\citeyear{kamath2018survey}) provide examples where given a natural language utterance and context 
in the form of a relational database, the system first converts the natural language utterance to a SQL query that is then executed 
using standard SQL grammar to retrieve the answer.
This is also similar in approach to how some teams that participated in the WikiSQL task \cite{zhongSeq2SQL2017} developed natural language interfaces for relational databases. 
However, writing and maintaining rules is not scalable especially as more complex types of reasoning become needed. 
The data annotation itself becomes hard to manage efficiently as more functionalities need to be supported. 
Furthermore, deep semantic parsing and reliably extracting attributes and relations and operating on multi-sentence input remains a challenge.

\paragraph{Satisfiability-based Approaches} \citeauthor{wang2019satnet} (\citeyear{wang2019satnet}) propose to integrate a differentiable maximum satisfiability solver into the loop of larger deep learning systems, 
and use this approach to successfully learn logical structures such as the rules of Sudoku. 
Previous works have shown that temporal reasoning can be modeled as a propositional satisfiability problem \cite{pham2008modelling}; however, generalizability to other types of reasoning needs further investigation. 
Although covering a rich class of problems, these approaches impose a structure on the reasoning problem \cite{clark2020transformers}, 
i.e., learning of logical structure specifically as expressed by satisfiability problems. 

\paragraph{Neuro-symbolic Approaches} Neuro-symbolic systems are hybrid models that leverage neural networks and symbolic reasoning to integrate learning and reasoning. 
\citeauthor{besold2017neural} (\citeyear{besold2017neural}) provide a survey of how symbolic approaches for reasoning are integrated with the machine learning approaches that bring in reasoning. 
More recently, \citeauthor{dong2019neural} (\citeyear{dong2019neural}) propose Neural Logic Machines and apply them to different tasks such as relational reasoning and sorting. 
\citeauthor{arabshahi2020conversational} (\citeyear{arabshahi2020conversational}) propose an end-to-end differentiable solution that uses a Prolog proof trace to learn rule embeddings from data, and apply their approach to the task of uncovering commonsense presumptions. 
Similarly, \citeauthor{xie2019embedding} (\citeyear{xie2019embedding}) generate a graph model to embed logic rules into the prediction. 
However, \citeauthor{ding2020object} (\citeyear{ding2020object}) show that a fully-learned neural network with the right inductive biases 
can outperform neuro-symbolic approaches in the context of spatio-temporal interactions between objects.

\paragraph{Transformer Approaches} \citeauthor{clark2020transformers} (\citeyear{clark2020transformers} ) and \citeauthor{talmor2020leap} (\citeyear{talmor2020leap}) propose to train transformers to reason over natural language sentences, 
bypassing a formal representation and show such reasoning over language is learnable. 
\citeauthor{ding2020object} (\citeyear{ding2020object}) apply a similar technique to visual question answering and 
show that their approach outperforms neuro-symbolic approaches. 
\citeauthor{han2020deer} (\citeyear{han2020deer}) use a similar approach to fine-tune a language model for event temporal reasoning. 
Our approach builds on top of these works in that we integrate reasoning into task-oriented dialogues and go beyond predicting a 
true/false response for an input and instead directly predict the answer when the model has the information or 
extract constraints when it has partial information.

\paragraph{Knowledge Grounding in Dialogue}
Similar to how \citeauthor{zhongSeq2SQL2017} (\citeyear{zhongSeq2SQL2017}) retrieve knowledge from Wikipedia, approaches such as  
\cite{ghazvininejad2018, neelakantan2019, gopalakrishnan2019}
retrieve knowledge from a database to be incorporated into dialogue. These approaches extend the seq2seq approach to condition on the facts present in the knowledge bases. While this is a promising architecture, such approaches are good for applications such as knowledge-grounded open domain chat but not for supporting reasoning in task-oriented dialogues. 

\paragraph{Other Approaches}
There are also other techniques in the literature such as integrating rules defined in first-order logic with knowledge distillation \cite{hu2016harnessing} 
that are outside the above categories. 
There have also been efforts such as CLUTRR \cite{sinha2019clutrr}, bAbI dataset \cite{weston2015towards}, Single Rule Test \cite{richardson2020probing}, QuaRTz dataset \cite{tafjord2019quartz}, HotpotQA \cite{yang2018hotpotqa}, and ROPES (Reasoning over Paragraph Effects in Situations) \cite{lin2019reasoning}, 
that focus on creating benchmarks for reasoning that measure how well existing systems perform on generalized reasoning.

\section{Problem Statement}

Task-oriented dialogue systems use a natural language understanding component to extract semantic meaning from the user utterance, and 
elicit constraints from users to understand their goals in order to provide information, 
perform a task or provide options and alternatives for users to choose from, retrieved from external knowledge sources (e.g, through API calls). 
As such, we focus on reasoning over tasks and recommended items in the dialogue which are typically characterized by different attributes, 
for example, movie names and show-times for a ticket booking scenario. 
These systems rely on such representations to answer user queries such as \emph{``At what time is Vertigo playing?''} by performing API calls 
(e.g. \emph{searchTime(movie=Vertigo)}) which return the required information in a structured form 
(Movie=Vertigo, Times=[12:30-2:30 PM, 3-5 PM], Theater=Cineplex). 
The required information is then returned to the user in natural language (e.g. \emph{Vertigo is playing today from 12.30 to 2.30 PM and from 3 to 5 PM}.). 
However, in most currently available task-oriented dialogue systems if the user said next ``Book me the earliest one,'' 
although this information is already available to the system from the previous API call, 
given the lack of reasoning abilities the system would either not support such queries, or it would have to make an additional independent API call 
(e.g., \emph{searchEarliestTime(movie=Vertigo)} or \emph{searchTime(movie=Vertigo, modifier=earliest)}), 
creating redundant latency in the response and requiring the developer of the system to add APIs/rules to handle these use cases.

Given the above description, our objective is to train a model to learn how to reason over the information provided in the context. 
We assume the following scenarios for each user utterance:  

{\bf 1. Reasoning-required, answer available in the context}: The case where the user utterance requires reasoning and it is possible to infer the answer to the user query from the information returned by the previous API calls (e.g., ``Give me the earliest one''). Rather than extracting mentions and querying the knowledge base again, in this case the model directly outputs the predicted next system action along with its arguments.

{\bf 2. Reasoning-required, answer not available in the context}: The case where the user utterance requires reasoning, but it is not possible to infer the answer to the user query from the information returned by the previous API calls (e.g., ``Show me cheaper options''). In this case the model extracts constraints from the user utterance to be passed to the back-end API.

{\bf 3. Reasoning-not-required}: The case where the user utterance does not require reasoning (e.g., ``Please repeat'').

In order to support these scenarios, the model needs to learn to 1) compare between different items based on numerical and categorical attributes, 
2) compare across a list of numerical values to identify the minimum/maximum value among alternatives, 3) be able to formulate constraints when it is not possible to infer the answer to the user query given the dialogue context but partial inference can be made, and 4) respond no answer when no reasoning is required for answering the user's request.

 \begin{figure}[t]
  \includegraphics[width=\linewidth]{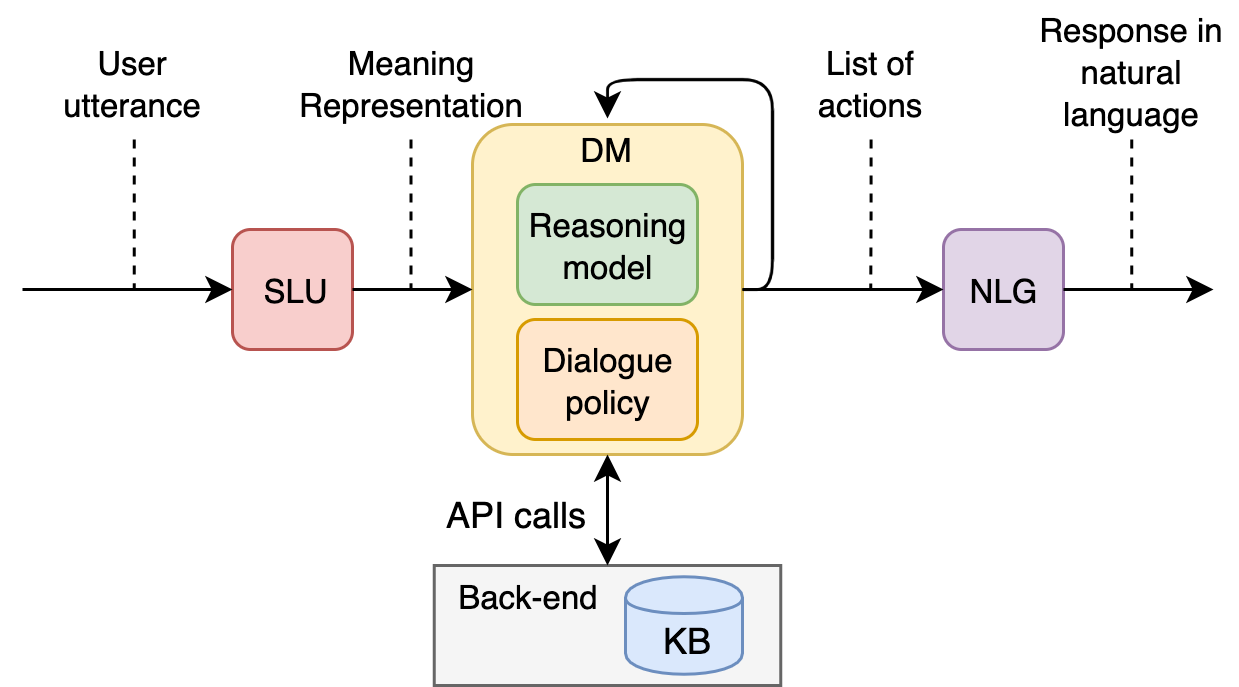}
  \caption{The reasoning model can be easily integrated in task-oriented dialogue architecture, as a component of the Dialogue Manager, i.e., the module in charge of predicting the next system action.}
    \vspace{-15pt}
  \label{fig:figure_0003}
\end{figure}

Figure \ref{fig:figure_0003} shows the overall architecture of a dialogue system with the reasoning model. 
The new model is part of the dialogue manager which predicts the next system action, 
along side a domain specific dialogue policy. 
The dialogue policy can predict API calls for retrieving information from a back-end Knowledge Base (KB) or can predict a list of natural language generation (NLG) actions for communicating information to the user (requesting constraints, informing available options, etc.). 
The reasoning model is added as a modular component that runs along-side the dialogue policy model. Although it would be possible to combine the two models, e.g, by extending the reasoning model to also predict domain specific APIs and actions, we believe that this modular architecture allows the reuse of a trained reasoning model across different domains and tasks. 

 \begin{figure*}[t]\centering
  \includegraphics[width=1.0\linewidth]{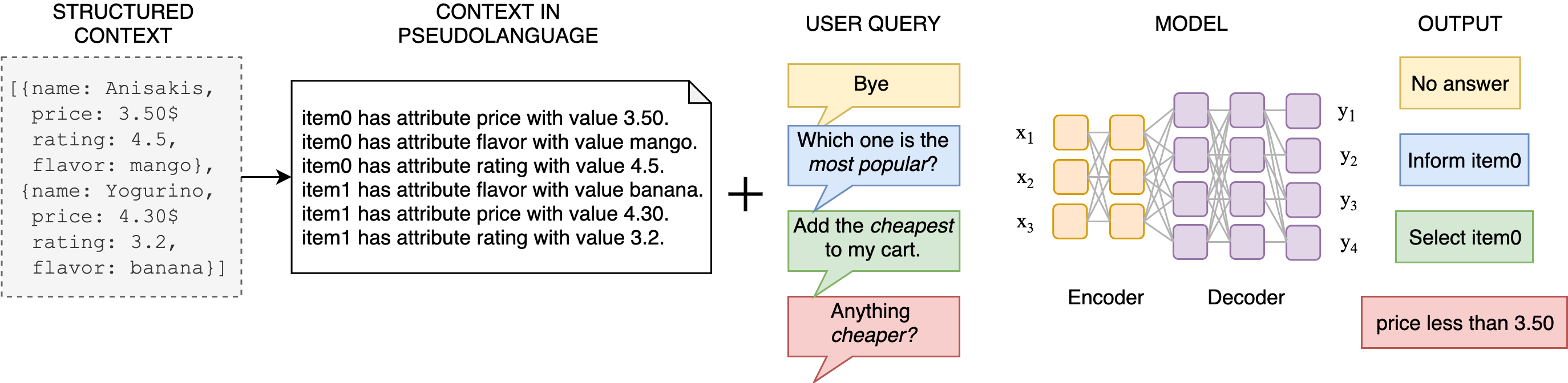}
  \caption{Task structure for the generative model.}
  \vspace{-15pt}
  \label{fig:figure_0004}
\end{figure*}

\section{Method}
In this work we propose to fine-tune transformers to learn logical reasoning over dialogue context in the form of natural language sentences, 
bypassing a formal representation and showing such reasoning over language is learnable. 

\subsection{Data Generation}
\label{sec:data-generation}
We describe a general methodology\footnote{Upon acceptance, we plan to release the data generated with the described methodology with the research community.} for automatically creating a dataset for logical reasoning in task-oriented dialogue systems. 
Each example in the dataset is a triple (user-query, context, answer), where the user-query refers to the last user utterance, 
the context refers to the dialogue context and information returned by API calls to the back-end system 
(see an example in Figure \ref{fig:figure_0001}), 
and the answer refers to the next action to be taken by the dialogue system. 
The user-query and the context constitute the information given as input to the model, while the answer represents the output. 

In order to simulate the context, the objects returned by API calls to the back-end system, we assume an available knowledge base (KB). 
We further assume that the KB will have different items, identified by an item-name (e.g., \emph{Yogurt Anisakis}), 
an item-type (e.g., \emph{yogurt}), and a series of attributes, each with an attribute key and value (e.g., \emph{price: $\$3.40$}). 
For generalizability, we do not assume that all item types have the same attributes, nor that all items of the same type have the same attributes. 

\begin{table*}[bp]
\centering
\resizebox{\textwidth}{!}{
\begin{tabular}{llll}
\hline
\textbf{Type} & \textbf{Name} & \textbf{Template} & \textbf{Example}\\
\hline
Context Statement & IsA & [subject] is [object]. &  Yogurt Anisakis is a yogurt. \\
Context Statement & HasAttribute & [subject] has attribute [attribute] with value [value]. & Yogurt Anisakis has attribute price with value $3.55$. \\
\hline
User Query & Inform  & I want something [predicate] [value]. & I want something cheaper than $\$5$. \\
& & Which one is [predicate]? & Which one is the cheapest? \\
User Query & Inform\_TrueFalse & Is [subject] [predicate]? &  Is Yogurt Anisakis the cheapest?\\
User Query & Select & Select [predicate]. & Select the cheapest. \\
\hline
\end{tabular}}
\caption{\label{table:Table_0001}
Examples of templates for context statements (in pseudo-language) and user queries (in natural language)
}
\end{table*}

The data generation procedure consists of four main steps:

\noindent
{\bf 1. Items sampling}: In order to construct input-output pairs for training, we first randomly select $k$ items, where $0 \le k \le k_{max}$, 
with the same item-type to create the input context $c$. 
While in this work we compare items of the same item-type, this is not a strict requirement of data generation. 
The motivation behind this choice is given by a typical scenario of a task-oriented dialogue system where a user might search for a specific object 
(movie times of Vertigo) 
and the system would subsequently present different options for that object 
(``Vertigo is playing today from 12:30 to 2:30 PM and from 3 to 5 PM.'').

\noindent
{\bf 2. Context conversion to pseudo-language}: Once a set of items has been sampled, we transform the structured information (list of triplets) 
associated to each item into pseudo-language by using a template-based approach, as in Figure \ref{fig:figure_0004}. 
Our templates are constructed in a domain-agnostic way, so that they would be directly applicable to other scenarios. 
We define two main types of statements in pseudo-language, each one associated to a specific template (see first two rows in Table \ref{table:Table_0001}). 
The \texttt{IsA} template is used to define the type of an item, while the \texttt{HasAttribute} relation is used for triplets expressing the value of a given attribute for the specified item. 
We note that other templates for the context statements could easily be created to accommodate different scenarios. 
Finally, we concatenate all the generated statements, after randomizing their order for improving robustness, to form the final input context. 


\begin{table*}[t]
\begin{adjustbox}{width=1.6\columnwidth,center}
\begin{tabular}{lllll}
\hline
\textbf{Reasoning} & \textbf{Answer} & \textbf{Action} & \textbf{Example} & \textbf{Output}\\
\textbf{Required} & \textbf{in Context} & \textbf{Type} & & \\
\hline
Yes & Yes & Inform & Is the first one cheaper than the second one? & inform <true/false> \\
Yes & Yes & Inform & Which one is the cheapest? & inform <item\_name> \\
Yes & Yes & Select &  Add the cheapest to my cart. & select <item\_name> \\
Yes & No & Constraint & Give me something cheaper & <relation> <attribute> <value>  \\
No & $-$ & No Answer &  Find yogurt. & NoAnswer \\
\hline
\end{tabular}
\end{adjustbox}
\caption{\label{table:Table_0002}
Output space. In cases where there are multiple answers/constraints, they are concatenated with \texttt{and}.
}
\end{table*}

\begin{table}
\begin{adjustbox}{width=0.8\columnwidth,center}
\begin{tabular}{ll}
\hline
\textbf{User Utterance} & \textbf{Constraint} \\
\hline
Give me something vegan. & include diet vegan \\
I don't want mango. & exclude flavor mango \\
It should cost $\$1.50$. & equal price $1.50$ \\
I want it cheaper than $\$2$ & less-than price $2$ \\
Anything more popular? & more-than rating $4.5$ \\
\hline
\end{tabular}
\end{adjustbox}
\caption{\label{table:Table_0003}
Examples of constraints representation, given as context the one in Figure \ref{fig:figure_0003}.
}
\end{table}

\noindent
{\bf 3. Query generation:} In this step we generate a set of user queries $q$ suitable for the given context using templates, 
thus generating several number of different input pairs $(c, q_{i})$ where $i$ is an index over possible queries related to the context $c$. 
Note that templates for the queries are manually created for each attribute, but they are all agnostic from the domain of the task-oriented dialogue system. 
Examples of user queries are shown in Table \ref{table:Table_0001}. 
As it can be seen, each template for the user query was associated to the expected output action predicted by the system and the particular reasoning ability involved (e.g., \texttt{Inform}). 
We also consider more complex cases such as \emph{negation}, e.g., ``I don't want anything vegan,'' and \emph{conjunction}, 
e.g., ``Which is the cheapest one and doesn't have strawberry?''. 
Additionally, each template is associated with several different surface form variations to add robustness to the model. 
Each generated user query is then prepended to the context $c$. 
An additional optional post-processing step consists of converting all the numerical values in the user queries from written to spoken format (e.g. \emph{``$\$3.50$''} is converted to \emph{``three dollars fifty''}). 
This step might be required in the context of a spoken dialogue system scenario, which takes directly as input the output of the Automatic Speech Recognition model. 

\noindent
{\bf 4. Output creation:} In the final step, for each generated input, we automatically create the output by combining the information from each template in regards to the action type to take and calculating the correct answer from the context, e.g., \emph{Yogurt Anisakis is the cheapest}. 
The output space consists of four main outcomes, as shown in Table \ref{table:Table_0002}, 
depending on whether reasoning is required to respond to the user utterance, and whether the answer is retrievable from the available context. 
We use the special token \texttt{NoAnswer} for user queries that do not require reasoning.
When the answer is retrievable from the context and reasoning is required, we further distinguish between two main cases: 
\texttt{inform}, when the user is simply seeking information (e.g., \emph{``Which one is the cheapest?''}), thus performing an Information-Transfer type of Dialogue Act (see \citet{bunt2010iso}), and 
\texttt{select}, when the user is requesting the system to perform a specific  action (e.g., \emph{``Add the cheapest to my cart.''}), an Action-Discussion Dialogue Act. 
For the inform action, we also distinguish in the output space between True/False questions 
and open-answer questions. 


In the case of constraint extraction answers, i.e., when the user utterance requires reasoning but the context has partial information, 
the output consists of the list of constraints extracted from the user query and concatenated with \texttt{and}, as shown in Table \ref{table:Table_0003}. 
The constraints extracted from the user query depend on the context, not only in terms of action to take (whether to provide an answer directly or to extract constraints), but also in terms of constraints generation. 
In the last row of Table \ref{table:Table_0003}, for user query ("..more popular?") the reasoning model relies on the context by looking at the ratings of the available products to extract the appropriate rating constraint (e.g, more-than rating $4.5$).




\subsection{Training Procedure}



In order to teach the model rules such as inverse relations and transitivity by example, 
we investigate the use of appending to the context clues that describe the relations of one or more items. 
These clues are appended to the final input context during training, but not at inference time. 
We consider two types of clues: 1) 
\emph{Comparative clue} describes a comparison of two items in the context along a specific attribute. 
The template for this clue is: [\texttt{subject}] is [\texttt{predicate}] [\texttt{object}], 
where \texttt{predicate} refers to the quality regarding which the items are being judged (e.g., ``cheaper than'', ``pricier than'', ``less than'', ``equal to''). 
2) \emph{Superlative clue} describes an object at the upper/lowest range of a specific attribute. 
The template for this clue is: [\texttt{subject}] is [\texttt{predicate}] with value [\texttt{value}].
Using the base data generation and clue generation, we are able to construct three types of training scenarios, as follows:

{\bf Case I - Clueless context:} This scenario uses the base context encompassing the information about the items' different attributes. 
This is also the scenario we expect at inference time. 

{\bf Case II - Comparative clues:} In this scenario, we sort the items in the base context according to the values of their attributes and append to the base context the comparative relation between pairs of items that are neighbors. 
The direction of comparison selected is random (e.g. ``A is larger than B'' or ``B is smaller than A'') and independent from the user query.
This scenario is designed to assess the ability of the model to learn \textit{inverse} relations, since in some queries users will ask for a relation in the opposite direction in regards to the comparative clue in the context (e.g., user asks ``Is the second one cheaper than the first one?'' while in the context we have ``A is pricier than B''), so that the model could learn that these two statements are equivalent. 
When we have more than two items in context, we can also assess the ability of the model to learn \textit{transitivity}, as we might have cases where the user asks ``Is the first one pricier than the third one?'' and in the context we have ``A is pricier than B'' and ``B is pricier than C''.

{\bf Case III - Superlative clues:} In this scenario, besides comparative clues, we also add superlative clues to the context to give hints to the model about which item in the context has the extreme value of the attributes (e.g. ``A is the cheapest''). 

We pick the number of items in each context randomly from $0$ to $k_{max}$, so that the model can be robust in its prediction for different number of items in the context. 
We also consider an additional training procedure, which we refer to as Case IV, where we randomly select one of Case I, Case II, or Case III as our context. 
The random selection of context helps the model to experience all three different cases and by cross learning between different cases, 
it learns to apply the inverse and transitivity rules for examples with Case I context to draw the right conclusion.

\section{Experiments}
\label{sec:label_experiments}

We showcase our proposed methodology in the context of a dialogue system for a shopping assistant (see Appendix \ref{sec:appendixA} for an example interaction). 
We use an ontology for data generation which consists of \texttt{item-type} (e.g. yogurt) and \texttt{item-name} 
(``Greek yogurt Anisakis'') and each item is characterized by two numerical attributes \texttt{price} and \texttt{rating}, 
and two categorical attributes \texttt{diet} and \texttt{flavor}. 
This choice of attributes can help us explore and assess the model's performance based on attribute's characteristics. 
Table \ref{table:statistics} summarizes the size of the catalog or range of values for each attribute.

\begin{table}
\centering
\begin{adjustbox}{width=0.7\columnwidth,center}
 \begin{tabular}{p{15mm} p{18mm} c  c} 
 \hline
 {\bf Rating} & {\bf Price} & {\bf Diet} & {\bf Flavor} \\ 
 \hline
 Bounded \newline Numeric & Unbounded \newline Numeric & 10 & 10K \\ 
 \hline
\end{tabular}
\end{adjustbox}
\caption{\label{table:statistics} Attributes and their catalogs size. }
\end{table}
We consider two settings for assessing the logical reasoning capability of transformer models. 
In the first setting, we fine-tune RoBERTa-base \cite{roberta} with a training dataset generated for reasoning using only numerical attributes. 
In this setting, we only focus on True/False prediction for each query $q$ given the facts provided in the context $c$. 
The objective of this experiment is to understand whether transformer models can learn to reason over numerical attributes. 
In the second setting, we use a T5 model \cite{t5} fine-tuned for the UnifiedQA data \cite{unifiedqa}, 
to predict a sequence similar to one given in Table \ref{table:Table_0002}. In both cases, we use disjoint catalogs to generate examples for train/dev/test datasets to avoid over-fitting to attribute values.


\subsection{True/False Queries}
We consider True/False reasoning over attributes such as assessing a conclusion about the comparison of two values of an attribute, or
finding minimum or maximum value among list of values of an attribute for several items.
Example queries include ``is the second item the cheapest one'' and ``is the first one cheaper than the fourth one''. 
We fine-tune RoBERTa to predict True/False for each ($q$, $c$) by adding a classification layer on top of the RoBERTa encoder model to perform binary classification. 
The training hyper-parameters for fine-tuning this model are provided in Appendix \ref{sec:appendix_parameters}. 
For these experiments, we generate $120$K samples for train, $5$K for dev, and $25$K for test set.

{\bf Clueless Training}: In this case, we only add \texttt{IsA} and \texttt{HasAttribute} relations and don't include any clue in the context $c$ in the training data (i.e., Case I). 
For each generated context, the data generation process attaches all possible forms of queries and the potential true/false label and adds them to training samples. For evaluation, we generate the test samples in a similar fashion. 
Table \ref{tab:clueless} summarizes the model performance for predicting the right label for each query given the context with $k \in {2, 3, 5}$ number of items in the context. 
\begin{table}
\begin{adjustbox}{width=0.8\columnwidth,center}
 \begin{tabular}{c C{12mm} C{12mm} C{12mm} } 
 \hline
\textbf{Train/Test} & \textbf{I/I} & \textbf{II/II} & \textbf{III/III} \\ 
 \hline 
 2 items & $90\%$ & $97\%$ & $97\%$  \\ 
 3 items & $88\%$ & $95\%$ & $95\%$ \\ 
 5 items & $77\%$ & $91\%$ &  $93\%$  \\ 
 \hline
\end{tabular}
\end{adjustbox}
\caption{\label{tab:clueless}Roberta-Base model performance for T/F Reasoning over Price and Rating.}
\end{table}
We can see that by increasing the context size (or number of returning items from back-end) the model performance decreases. 
To understand how well a model with larger $k$ with comparative or superlative clues can generalize to fewer number of items in context, 
Table \ref{tab:clueless2} shows the performance of a model trained with context size of $5$ items using Case II or Case III samples and tested on samples generated by Case I and with $k \in {2, 3, 5}$ items. 
We observe that the model does not generalize to different context sizes if we fix the number of items in the context during model training. 
\begin{table}
\begin{adjustbox}{width=0.8\columnwidth,center}
 \begin{tabular}{p{25mm} C{18mm} C{18mm}} 
 \hline
 \textbf{Train} $\rightarrow$ &  \textbf{Case II} & \textbf{Case III} \\ 
 \textbf{Test}   $\downarrow$ & \textbf{(5 items)}  &  \textbf{(5 items)}\\ 
 \hline 
 Case I, (2 items)& $75\%$ &  $76\%$\\ 
 \hline
 Case I, (3 items) &  $70\%$ & $71\%$ \\ 
 \hline
 Case I, (5 items) &  $67\%$  &   $69\%$ \\ 
 \hline
\end{tabular}
\end{adjustbox}
\caption{\label{tab:clueless2} Train on Case II or Case III with $5$ items in all the contexts and test on Case I with $2$, $3$, or $5$ items.}
\end{table}

{\bf Clue-Aware Training}: To resolve the issues in clueless training, we add comparative and superlative clues randomly to each context during the training such that the model can learn the inverse and transitivity rules; 
and also we add random number of items to each individual context (up to $k_{max}$). 
Note that we do not add clues to the context during evaluation/inference. 
Results in Table \ref{tab:clueAware} show the accuracy performance of models trained using samples generated by Case IV and tested on Case I (clue-less), Case II (only comparative clues), and Case III (both comparative and superlative clues) samples.
\begin{table}
\begin{adjustbox}{width=0.8\columnwidth,center}
 \begin{tabular}{c C{12mm} C{12mm} C{12mm} } 
 \hline
\textbf{Train/Test} & \textbf{IV/I} & \textbf{IV/II} & \textbf{IV/III}  \\ 
 \hline 
up-to 5 items & 98.70\% & 99.70\% & 99.70\%  \\ 
 \hline
\end{tabular}
\end{adjustbox}
\caption{\label{tab:clueAware}Training with CaseIV: Roberta model performance for T/F reasoning over numerical attributes.}
\end{table}
From the results, we observed that adding clues during model training helps the model to achieve better performance. 

\subsection{Beyond True/False Queries}
For this set of experiments, we pick the T5 transformer model which can enable us to perform text-to-text prediction. 
Similar to \cite{unifiedqa}, we remove the task prefix that has been used in the original T5 models, since we will use this model only for a single reasoning task within our defined framework. 
To take advantage of transfer learning from other publicly available question-answering datasets, 
we start our fine-tuning from the pretrained Unified-QA-T5 small model. 
We generate $100$K samples for training dataset, $5$K for dev, and $20$K examples for each test set. 
In our test set we make sure that for each element in Table \ref{tab:T5-model}, we have at least $5$K examples. Samples are generated as described in Section \ref{sec:data-generation}.
The training hyper-parameters for fine-tuning this model are provided in Appendix \ref{sec:appendix_parameters}.

In Table \ref{tab:T5-model}, we summarize the performance of the fined-tuned model for different scenarios, 
reporting the results separately for pair of ($q$, $c$) such that  $q$ can have one (e.g., ``Give me something organic') or two attributes (e.g., `Something cheaper than $\$100$ but not vegan'') about user-preferences. 
We use the exact-match (EM) accuracy metric to evaluate model performance.
We can observe that the model can achieve an EM accuracy of over $90\%$ across all the scenarios. 
Furthermore, we see that when increasing the number of items in the reasoning context, predicting the correct Inform/Select or Extract output form becomes harder with more attributes in the user query.   
Evaluating the model performance on all examples (about $8$K samples) from our test set that include spoken form of numerical values in $q$ (e.g., ``Give me something cheaper than five dollars''), we observe $95\%$ EM accuracy, showing the ability of the model to compare written form and spoken form versions of numbers. We should note that the accuracy of the model for predicting the cases with no reasoning (e.g., ``Checkout please'') is important because it makes the integration with the overall dialogue system simpler where the model can delegate to the domain specific dialogue policy. 
In our experiments, we observe an accuracy of $100\%$ on these cases; however, this value can vary by increasing the size of out-of-domain space/vocabulary. 

\begin{table}
\begin{adjustbox}{width=0.8\columnwidth,center}
 \begin{tabular}{ p{15mm} p{5mm} | p{20mm}  p{20mm} } 
 \hline
\# of Attr.s  & $k_{m}$ &   Inform/Select & Extract  \\ 
 \hline 
\multirow{5}{4em}{1} & 0 &  --                            &  $99.5\pm 0.02\%$      \\ 
								& 1 &   $98.6\pm0.05\%$  &  $99.2\pm0.03\%$     \\ 
                                    & 2 &  $97.3\pm0.05\%$   &  $98.5\pm0.05\%$     \\ 
                                    & 3 &  $97.0\pm0.05\%$   &  $98.0\pm0.03\%$     \\ 
                                    & 4 &  $96.0\pm0.10\%$  &  $98.0\pm0.05\%$     \\ 
                                    & 5 &  $95.5\pm0.09\%$  &  $96.0\pm0.06\%$     \\ 

 \hline
\multirow{5}{4em}{2} &0 &   -- 						    &  $98.6\pm0.03\%$     \\ 
								& 1 &   $98.5\pm0.05\%$   &  $97.8\pm0.02\%$  \\ 
                                    & 2 &  $95.0\pm0.08\%$   &  $96.7\pm0.01\%$  \\ 
                                    & 3 &  $94.5\pm0.05\%$   &  $96.3\pm0.03\%$  \\ 
                                    & 4 &  $91.5\pm0.09\%$   &  $95.0\pm0.03\%$   \\ 
                                    & 5 &  $90.0\pm0.11\%$   &  $93.5\pm0.06\%$   \\ 

 \hline
\end{tabular}
\end{adjustbox}
\caption{\label{tab:T5-model}EM accuracy for test sets with different number of attributes, context size, and reasoning task.}
\end{table}

\section{Conclusions} 

In this paper, we proposed an architecture for the integration of a reasoning model in task-oriented dialogue systems. 
We formulated the problem as a sequence prediction problem given a user query and context, and presented an approach for generating data and 
fine-tuning generative models to reason over a set of facts in the dialogue context. 
We demonstrated our approach for a shopping assistant and reported experimental results for different formulations of the problem.
We showed that these models can learn to do logical reasoning to 1) answer questions form the dialogue context when all the information is available, 
2) extract constraints when partial information is available, and 3) delegate to the dialogue policy when no reasoning is required.
For future work, we plan to investigate the application of our method to other reasoning tasks (e.g., temporal and spatial reasoning). 
Furthermore, we plan to explore how logical reasoning can be used to disambiguate with the user when multiple conclusions can be made.

\bibliographystyle{acl_natbib}
\bibliography{custom}

\clearpage

\newpage
\newpage

\appendix

\section{Example Interaction with the Shopping Assistant}
\label{sec:appendixA}

The following is an example interaction with the shopping assistant with our reasoning model integrated with the dialogue policy.

\begin{table}[htbp]
\begin{adjustbox}{width=1.0\columnwidth,center}
 \begin{tabular}{ll} 
 \hline
 \textbf{Speaker}  &  \textbf{Utterance}  \\ 
 \hline 
User & I want to buy apples \\
Agent & The top 3 relevant products are: Organic \\
& Honycrisp Apple, Organic Gala Apple, and \\
& Organic Pink Lady Apple. Which one would you  \\
& like to add to the cart?\\
User& Which one is the cheapest?\\
Agent & The second one, Organic Gala Apple with a  \\
& price of \$2.49.\\
User&Which one is the most expensive?\\
Agent &The first one, Organic Honeycripst Apple with a  \\
& price of \$3.99\\
User& Is the first one cheaper than the second one?\\
Agent &No. The first one, Organic Honeycripst Apple, \\
& costs \$3.99 but the second one, Organic Gala \\
& Apple, costs \$2.49.\\
User&Is the second one cheaper?\\
Agent &Yes, that is correct. The second one, Organic Gala \\
& Apple, costs \$2.49.\\
User&Give me something cheaper\\
Agent &Top relevant product is Fuji Apple. Would you \\
& like to add it to the cart?\\
User&Does it cost less than two dollars?\\
Agent &Yes, that is correct. Fuji Apple costs \$1.89.\\
User&Add it to my cart\\
Agent& I have added Fuji Apple to your cart. What else \\
& would you like to buy?\\
User& Buy grapes \\
Agent & The top 2 relevant products are: Red Seedless \\
& Grapes and Conventional Cut Grapes. Which one \\
& would you like to add to the cart?\\
User & Is the second one more popular? \\
Agent & No. The second one, Conventional Cut Grapes \\
& has a rating of 4.3 but the first one, Red Seedless \\
& Grapes has a rating of 4.5. \\
User & Add the highest rated one to my cart \\
Agent & I have added Red Seedless Grapes to your cart. \\
& What else would you like to buy? \\
User & Find cheese \\
Agent & The top 2 relevant products are: Zola and Muller. \\
& Which one would you like to add to the cart?\\
User & Anything vegan and cheaper than five \\
& dollars? \\
Agent & I couldn't find anything vegan with price less than \\
& five dollars. But I found Vegan Artisan cheese with \\
& price of \$7.99. Would you like to add it to the cart? \\
User & Sure \\
Agent & I have added Vegan Artisan cheese to your cart. \\
& What else would you like to buy?\\
User & What's in my cart? \\
Agent & You have 3 items in the cart with total amount \\
& of \$12.87. Do you want to hear detailed description? \\
User & Checkout \\
Agent & I have placed your order. \\
 \hline
\end{tabular}
\end{adjustbox}
\end{table}

\section{Training Hyper-parameters}
\label{sec:appendix_parameters}
In this section, we provide the parameters that are used to fine-tune the transformer models in this work. 
The following table summarizes the key parameters that are used during the fine-tuning of Roberta-base and UnifiedQA-T5-small pretrained models.
For the optimizer, we use AdamW \cite{loshchilov2017decoupled}.

\begin{table}[!h]
\begin{adjustbox}{width=\columnwidth,center}
 \begin{tabular}{c C{10mm} C{10mm} C{15mm} C{15mm} C{10mm} } 
 \hline
Model & Batch Size & Epoch & Optimizer & Learning Rate & Weight Decay \\ 
 \hline 
 Roberta-Base &  192  &  15  & AdamW & $10^{-5}$  &  0.1  \\ 
 \hline
UnifiedQA-T5-small  &  1024  &  20  & AdamW & $10^{-5}$  &  0.0  \\ 
 \hline
\end{tabular}
\end{adjustbox}
\end{table}

\section{Discussion}
\label{sec:appendix_discussion}
\subsection{Generalization to non-synthesis datasets}
\label{sec:appendix_discussion-non-synthesis}
One of the directions that currently we are working on is to create realistic (human based) conversations with logical reasoning use cases during the interactions with the dialog systems. This type of dataset can help us to evaluate the proposed idea with higher degree of confidence. Since no matter how much one spends time on generating synthetic datasets, there will always be some uncontrolled structures introduced by design of data simulation mechanisms that can corrupt the fair evaluation of deep neural network models and their learning process. However, we believe the True/False scenarios in our current study are less prone to this type of issues and are quite helpful in understating of reasoning capabilities such as negation, numerical comparison, or inclusion/exclusion of categorical values of our proposed algorithm, since model needs to learn the reasoning procedure. In other words, the only way to come up with the right prediction by model is to apply the underlying reasoning procedure to formulate the output True/False results. We will consider: a) better algorithms for generating training data, and b) more realistic general purpose possibly human in the loop training data to make the data generation more general and less domain specific, for future exploration.

\subsection{Error Analysis}
\label{sec:appendix_discussion-error-analysis}
During our evaluation, we observed that the Transformer models (such as Roberta and T5) performance degrades when the length of the reasoning context increases, i.e., the number of items in the context for reasoning are longer. Also based on the results on Table \ref{tab:T5-model}, we see that increasing the number of items in reasoning context leads to performance degradation. 
Another issue with Transformer models or in general LM models is during the output generation process beyond the True/False scenario. When the size of the output sequence length increases, e.g., there are several items that all satisfy the user-query. The prediction misses some of the items in the response after the length of the output sequence (number of predicted tokens/words) meets some threshold. This issue is related to both long sequence generation of LM models and also reasoning ability when the multiple items match the user-query's criteria which mostly occurs when the number of items in context are larger.

\subsection{Generalization to unseen attribute with common values}
\label{sec:appendix_discussion-unseen-attributes}
One of the aspect that we like to understand is the scalability/generalization of the proposed trained reasoning model to unseen attributes during the test time. There are two possibility for a new attribute: (1) doesn't shares values and keywords that user may use to describe the attribute compared to the attributes that are used during the training process e.g., \texttt{color} attribute for experiment in Section \ref{sec:label_experiments} \footnote{For query about the \texttt{color} user may use keywords such as:  [darker, lighter, warmer, red, blue, ..., etc. ] one, and attribute values are red, blue, dark blue, .... etc. which doesn't overlap with none of the attributes that we have already in our training dataset, i.e., \texttt{diet, flavor, price, and rating}}.  (2) shares same values but keywords that user may use to describe the attribute doesn't overlap with any of the ones used during the training process, e.g., \texttt{calorie} \footnote{For query about the \texttt{calories} user may use keywords such as:  [healthier, higher calories, more energetic..., etc. ] one, and attribute values are numeric value that are shared possibly with \texttt{price} and \texttt{rating} [considering we have done unit normalization for attributes]}. It would be very challenging to teach model in a few-shot manner to learn about attributes from bucket (1). However, based on our initial experiments we have seen that model can easily generalize to the attributes from bucket (2), by fine-tuning to small number of examples in a few-shot manner. For example, we fine-tuned the model which only trained for \texttt{diet, flavor, price, and rating} attributes and fine-tuned using only $100$ new reasoning context examples which had $calorie$ attribute as well. Table \ref{tab:calorie} summarize the model performance before and after fine-tuning.  The test set used for this analysis only has user-query about calories and includes  $3 K$ examples about Calorie attribute.

\begin{table}[!t]
\begin{adjustbox}{width=0.8\columnwidth,center}
 \begin{tabular}{c C{25mm}  } 
 \hline
\textbf{Model} &  EM accuracy \\ 
 \hline 
Before fine-tuning &  $33\%$   \\ 
After fine-tuning &   $80\%$  \\ 
 \hline
\end{tabular}
\end{adjustbox}
\caption{\label{tab:calorie} Model EM accuracy performance  before/after fine-tuning to new attribute \texttt{calorie}.}
\end{table}

\end{document}